\documentclass[letterpaper, 10 pt, conference]{ieeeconf}  

\IEEEoverridecommandlockouts                              

\usepackage{graphics} 
\usepackage{epsfig} 
\usepackage{cite}
\usepackage{amsmath} 
\usepackage[linesnumbered,ruled]{algorithm2e}

\DeclareMathOperator*{\argmin}{arg\,min}

\title{\LARGE \bf
An Untethered Brittle Star-Inspired Soft Robot for Closed-Loop Underwater Locomotion 
}

\author{Zach J. Patterson$^{1*}$, Andrew P. Sabelhaus$^{1}$ , Keene Chin$^{2}$, Tess Hellebrekers$^{2}$  and Carmel Majidi$^{1}$$^{2}$
\thanks{}
\thanks{$^{1}$ Z.J. Patterson,  A.P. Sabelhaus, and C. Majidi are with the Department of Mechanical Engineering, Carnegie Mellon University, 
            Pittsburgh, PA 15213, USA
        {\tt\small \{zpatters, asabelha, cmajidi\} @andrew.cmu.edu}. *Corresponding Author.}%
\thanks{$^{2}$ K. Chin, T. Hellebrekers, and C. Majidi are with The Robotics Institute, Carnegie Mellon University, Pittsburgh PA 15213, USA \tt\small \{keenec, tessh\} @andrew.cmu.edu}%
}

\begin{document}
\bstctlcite{MyBSTcontrol}

\maketitle
\thispagestyle{empty}
\pagestyle{empty}

\begin{abstract}
Soft robots are capable of inherently safer interactions with their environment than rigid robots since they can mechanically deform in response to unanticipated stimuli. However, their complex mechanics can make planning and control difficult, particularly with tasks such as locomotion. In this work, we present a mobile and untethered underwater crawling soft robot, PATRICK, paired with a testbed that demonstrates closed-loop locomotion planning. PATRICK is inspired by the brittle star, with five flexible legs actuated by a total of 20 shape-memory alloy (SMA) wires, providing a rich variety of possible motions via its large input space. We propose a motion planning infrastructure based on a simple set of PATRICK's motion primitives, and provide experiments showing that the planner can command the robot to locomote to a goal state. These experiments contribute the first examples of closed-loop, state-space goal seeking of an underwater, untethered, soft crawling robot, and make progress towards full autonomy of soft mobile robotic systems.
\end{abstract}

\section{INTRODUCTION}
\begin{figure*}[thpb]
    \centering
    \includegraphics[width=0.9\textwidth]{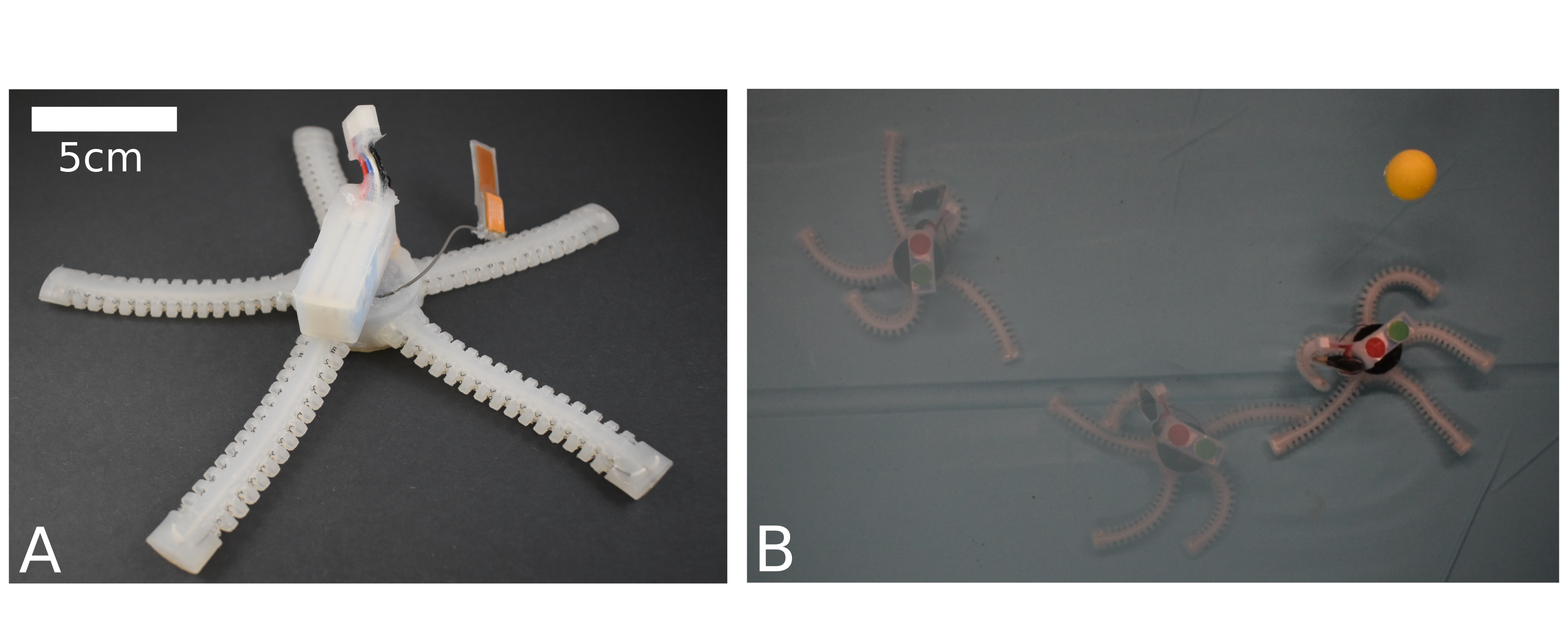}
    \caption{(A) The soft underwater untethered brittle star robot, PATRICK. (B) The robot, in its testbed, locomotes to a goal state (yellow ball).}
    \label{fig:robot_overview}
\end{figure*}

In recent years, roboticists have increasingly taken an interest in incorporating soft, flexible, and stretchable materials and structures into their designs \cite{Majidi2019}\cite{Rus2015}. By utilizing these types of materials and structures, soft roboticists hypothesize that robots can interact with their environments more safely and stably because of their inherent deformability - a concept often referred to as “morphological intelligence” \cite{Pfeifer2009}\cite{Paul2006}. However, robots must also be reliably controlled to accomplish useful tasks in the real world - a task made more difficult by the complexity of soft robotic systems. Soft robots make a trade off: by accepting a great deal of additional complexity and nonlinearity in their dynamics arising out of compliance, soft robots may be better able to perform with uncertainty, both in the environment and in their own state \cite{trivedi2008soft}.


For soft robots to be used more broadly in task-specific applications, it is necessary to create platforms that enable the investigation of techniques for performing specified tasks reliably on soft systems. Currently, soft robots are often designed to execute one of a few given functionalities and are better suited to ad-hoc demonstrations than to testing higher level controls, planning, and learning principles \cite{Huang2018}\cite{Lin2011}, with some notable exceptions \cite{Katzschmann2018}\cite{Calisti2015}. Groups such as Marchese et al. \cite{Marchese2015} have begun to integrate elements of autonomy from the literature on traditional robotic systems into the field of soft robot manipulation. This work is intended as an analogous step in the translation of those planning and control approaches to autonomous mobile soft robots.

We are not the first to address this need for more complex mobile soft robots. At the Robosoft Grand Challenge, participants competed to create soft robots capable of a rich variety of tasks \cite{Calisti2016}. OCTOPUS \cite{Calisti2011}, SUPERBALL \cite{Littlefield2019}, the cable driven quadruped by Bern et al. \cite{bern2019trajectory}, and Harvard's recent sea urchin robot \cite{paschal2019} are a non-exhaustive list of recent soft robots that are complex and high dimensional, in principle allowing exploration of interesting controls, planning, and learning concepts.

Challenges with soft robot control and modeling become especially acute when we place an untethered robot into an uncertain environment with its own power source and onboard electronics and ask it to perform with a high degree of autonomy. Relatively few soft robots in the literature have attained untethered functionality and even fewer have exhibited some degree of autonomy \cite{Katzschmann2018}\cite{Rich2018,Huang2018a,Mitchell2019,Tolley2014,Li2017}. To develop truly mobile and autonomous soft robots, we must be able to construct testbeds that allow us to begin to develop intelligent behaviors that exploit the potential of the systems. These testbeds should be untethered because (i) the incorporation of batteries and electronics is nontrivial and can have dramatic impact on robot design and (ii) a cable can have a significant effect on the locomotion of the robot, especially at the mass and length scales that most soft robots operate.

In this work, we seek to create a soft robotic platform that combines complex actuator dynamics and kinematic locomotion with a rich space of possible actions. The end result is a robot that we believe has a varied enough actuation space to explore questions of autonomy and control. See Fig. \ref{fig:robot_overview}A for an image of the robot. We begin this exploration with a feedback scheme enabled by a low-dimensional representation used for planning, leveraging empirical knowledge of the system dynamics within a reactive framework that allows the robot to follow a given goal.



\section{ROBOT DESIGN}


This work presents a set of designs for an untethered, underwater, crawling soft robot. We choose to focus on an aquatic environment because it is well suited to mobile soft robot design and applications. On the design front, the presence of the buoyancy force allows for more straightforward incorporation of batteries and electronics while maintaining locomotion capability. On applications, common fragility in ocean organisms and ecologies makes soft robots an appealing choice for noninvasive exploration and sampling \cite{Phillips2018}.


Our robot, named PATRICK, is inspired by the brittle star (\textit{ophiuroid}).
Brittle stars have five long and slender arms and are capable of a variety of large motions. 
Although they move relatively slowly in the grand scheme of nature, they are surprisingly agile and still capable of locomotion speeds around 2 centimeters per second \cite{Astley2012}. 
Most importantly, they fit our previously specified desire for a design with simple and kinematic limb motions that can lead to a large, rich space of potential action sequences. 
Their omnidirectionality also makes them an attractive source of inspiration since this enables greater agility and eliminates the need to expend time and energy for turning. 
The following section presents the design of the robot, including actuation and onboard electronics.

\subsection{Actuator Design}

To create limbs that function similarly to those of the brittle star, we chose shape memory alloy (SMA) springs (0.008in Dynalloy 90ºC Flexinol® with a 0.054in helical diameter) as an actuator because they are lightweight, easy to use, possess high work density, and can be powered and controlled directly with onboard electronics and a battery. Additionally, the water helps mitigate a core drawback of SMA by cooling it faster, allowing for higher frequency operation. The limb design was based on a design of Walters and McGoran \cite{Walters2011}. Each limb contains four SMA springs that are embedded in a molded silicone limb structure. The limb molds are printed with a Stratasys Objet 30 and the silicone is Smooth-On DragonSkin 10 NV silicone elastomer. The springs contract freely when current is applied and ribbed cutouts in the rubber allow the limb structure to bend as the springs contract (Fig \ref{fig:CAD_explosion}A). By actuating multiple at once or in sequence, the limbs are capable of reaching a large variety of states. PATRICK has five SMA-embedded silicone limbs that are 10 centimeters long with an elliptical cross section of 1.9 by 0.9 centimeters.

\subsection{Electronics}

\begin{figure*}[thpb]
    \centering
    \includegraphics[width=0.85\textwidth]{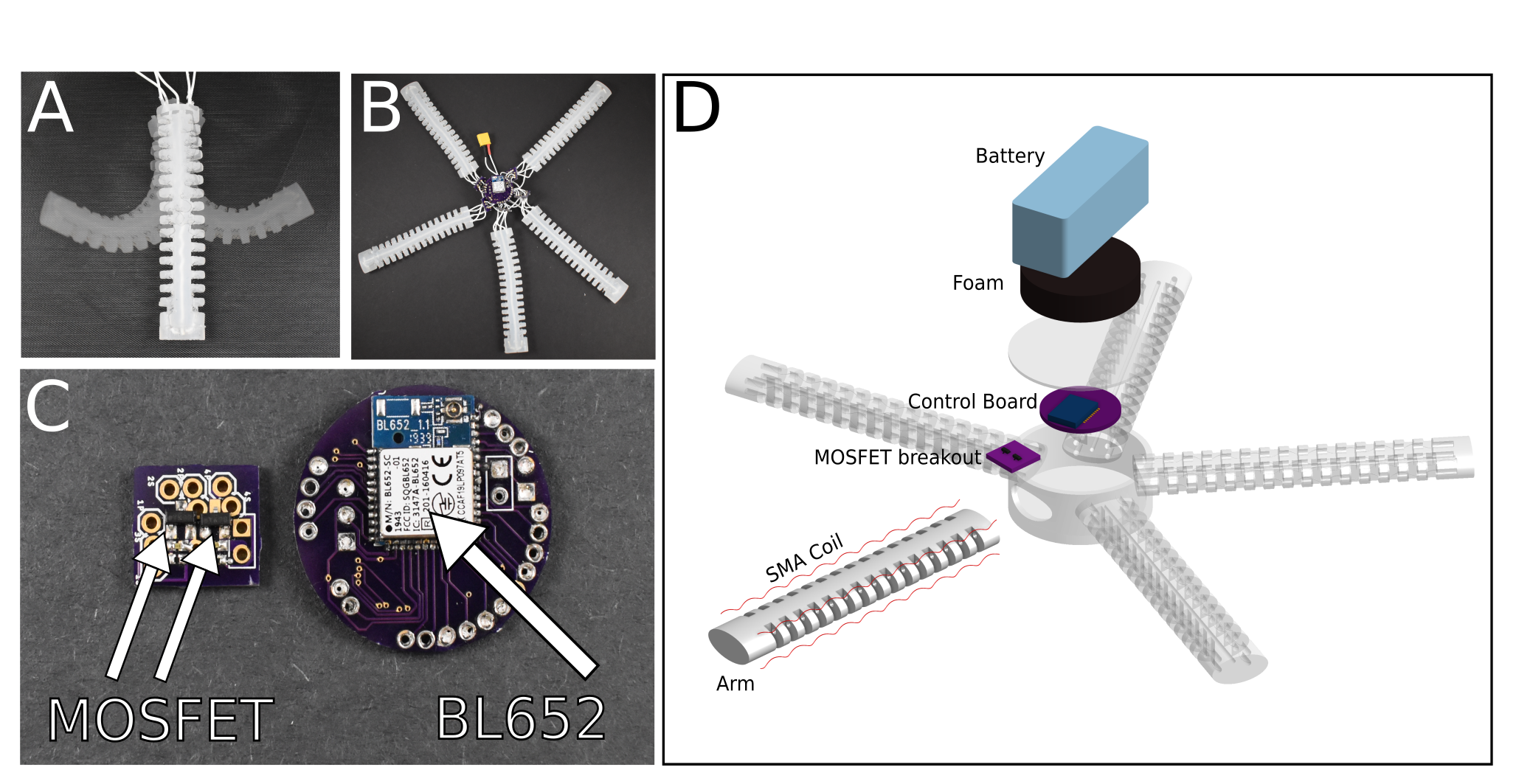}
    \caption{(A) Brittle star robot limb with SMA actuators. (B) Robot electronics and limbs before sealing into the central silicone hub. (C) Onboard system electronics including the microcontroller PCB and a MOSFET breakout board (5X). (D) System subcomponents from top to bottom: battery, foam for neutral buoyancy, control PCB, MOSFET breakout PCB, SMA spring, silicone arm.}
    \label{fig:CAD_explosion}
\end{figure*}

The robot contains power and control electronics within a central silicone hub that is sealed from water. It is controlled by a Laird BL652 module that contains a Bluetooth antenna and a nRF52832 microcontroller. The microcontroller communicates via bluetooth with an offboard nRF52832, which transmits control instructions from a computer via UART. Because the SMA limbs require high current to rapidly actuate, high powered MOSFETs are necessary to activate them with the microcontroller. We utilize SQ23 (40V) surface mount MOSFETs on breakout boards (Fig \ref{fig:CAD_explosion}C). For power, the robot uses a BETAFPV 11.1V, 300mAh, 45/75C drone battery.
To actuate one SMA wire, its corresponding MOSFET is pulled high and current flows through the wire, causing shape transition via Joule heating.

To construct the system, we wire the limbs to the MOSFET breakout boards which are then connected to the central control board as shown in Fig. \ref{fig:CAD_explosion}B. Finally, the electronics and limbs are cast into the central body of the robot, which is also made out of DragonSkin 10 silicone elastomer. The battery is separately sealed in silicone and placed on top of the robot, along with a cylinder of foam to get the robot close to neutral buoyancy. See Fig. \ref{fig:CAD_explosion}D for a full CAD explosion. The final diameter of the robot is 25 cm and the weight of the robot is 140 grams.

\section{ROBOT TESTBED}



\begin{figure*}[htbp]
    \centering
    \includegraphics[width=0.9\textwidth]{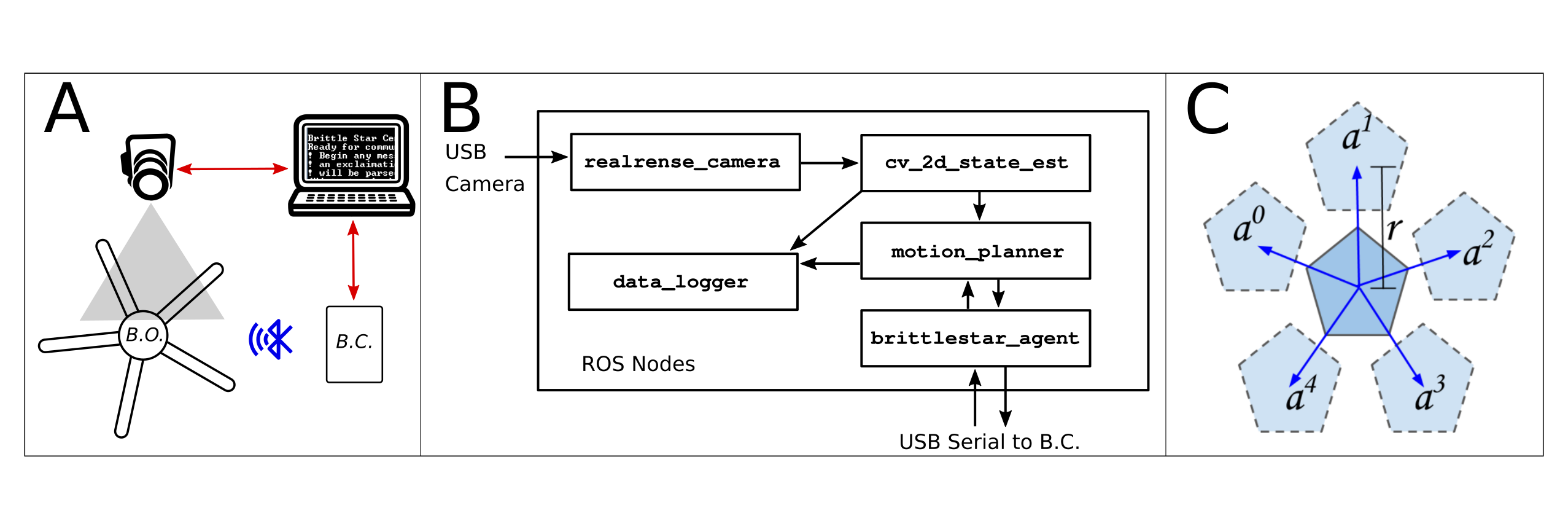}
    \caption{(A) System architecture for PATRICK and its testbed. A camera, pointed at the robot, tracks various markers via OpenCV running on a connected computer. Embedded software on the robot (``Brittlestar Onboard") communicates over Bluetooth Low Energy to similar software running on a separate board (``Brittlestar Central") for communication over USB (red) to various ROS nodes on the computer. (B) Architecture for the various nodes used in the Robotic Operating System (ROS) package for the PATRICK robot. The Intel RealSense camera package supplies frames, then four separate nodes perform state estimation from markers on the robot, motion planning, data logging, and communication with the robot itself. Commands are sent to and from the hardware robot via the \textit{brittlestar$\_$central} microcontroller attached over a USB serial port. (C)The idealized state transition model used by the planner.}
    \label{fig:sys_arch}
\end{figure*}

As part of this study, we created a testbed for the PATRICK robot to analyze different motions and provide closed-loop feedback.
To do so, three software subsystems were created (Fig. \ref{fig:sys_arch}A).
Embedded software on PATRICK (\textit{brittlestar$\_$onboard}) communicates with embedded software on a microcontroller attached to a computer (\textit{brittlestar$\_$central}).
A motion planning infrastructure built on the Robotic Operating System (ROS) and an offboard camera tracks markers on PATRICK for feedback.
All software is open-source and available online\footnote{https://github.com/softmachineslab/brittlestar}.


\subsection{Embedded Software}


The two microcontrollers involved in PATRICK's testbed are \textit{brittlestar$\_$onboard} and \textit{brittlestar$\_$central}, both running on the same Nordic nRF52832 system. 
The software for \textit{central} serves as a bridge between a USB serial port on the computer, sending and receiving strings of commands and responses over Bluetooth Low Energy with \textit{onboard}.

A command library was implemented that allows for activation and deactivation of each shape memory alloy (SMA) actuator on PATRICK.
The \textit{onboard} software operates each shape memory alloy actuator as a finite state machine.
SMA commands are sent to PATRICK in the form of (ON, OFF) for each of the 20 actuators.
By varying the amount of time each SMA is activated, the motion planning system can control the bending of each leg, in each direction.
Onboard safety timers trip after a timeout, deactivating each SMA if an OFF command has not been received within a time limit, preventing actuator burnout.

\subsection{ROS Infrastructure for Robot Tracking and Communication}

Alongside the flexible communication framework for PATRICK, an infrastructure of packages within the Robotic Operating System (ROS) was created that allow for visual tracking and executing motion plans.
Though the camera is offboard the robot, this is the first example of any high-level closed-loop locomotion control of a completely untethered underwater soft robot.

There are five ROS nodes running on the attached computer: camera frame capture, state estimation via OpenCV, motion planning, data logging, and communication with PATRICK via \textit{brittlestar$\_$central} (Fig. \ref{fig:sys_arch}B).
The camera, an Intel RealSense, has a manufacturer-supplied ROS node.
The remaining four are written specifically for PATRICK.


Here, we use a simple two-dimensional rigid body approximation for PATRICK, calculated by tracking two colored dots on the robot (Sec. \ref{sec:feedback_control}).
The position and orientation of its body are represented in 2D space based on the detection of these markers.
The motion$\_$planner node takes these state estimates, as well as a set of possible actions from the brittlestar$\_$agent (Sec. \ref{Sec:motion_primitives}), and selects one action to send to the agent.
The agent itself sends ON/OFF commands over USB serial to \textit{central} based on the desired action.


A test procedure consists of a camera calibration, then execution of a given motion plan.
Here, a second set of markers are placed on the floor of PATRICK's environment, and using them, a transformation is automatically calculated between the frame of the camera and the world frame.
\section{CLOSED-LOOP LOCOMOTION PLANNING}

To demonstrate proof-of-concept closed-loop locomotion of PATRICK within its testbed, the motion planning framework was implemented as a search over a set of hand-tuned motion primitives.
The following section gives the two-dimensional model that was used, the chosen representation of the motion primitives, as well as the greedy planning algorithm that was then used for hardware experiments in Sec. \ref{sec:results}.

\subsection{Motion Primitives}\label{Sec:motion_primitives}



PATRICK exhibits a rich set of dynamics based on its 20 actuators distributed across 5 limbs.
The underlying mechanics include both soft material deformation and thermal dynamics of shape-memory alloys, which are highly coupled and nonlinear.
This work instead chooses a low-dimensional representation of PATRICK's state and actuation space, representing actions as combinations of inputs, as is often used for computational efficiency in the search-based planning community \cite{likhachev2010search}.
The following section describes the model that was used, as well as the motion primitives which were discovered.
The inherent robustness of PATRICK's soft limbs allows for successful locomotion despite the low-dimensional model.

The two colored trackers on PATRICK allow for determining the position and orientation of its body in the global frame.
Together, with the SMA inputs at each timestep, the robot's model is

\[
\mathbf{x}_t=[x, y, \theta], \quad \quad \mathbf{u}_t \in \{0,1\}^{20},
\]

\noindent where each of the 20 SMAs is activated or deactivated.
The goal state of the robot is represented in-plane as $\bar{\mathbf{x}} = [\bar{x}, \bar{y}]$, designated by a third colored marker in the test environment.
The cost functions in Sec. \ref{sec:feedback_control} do not currently weigh the $\theta$ state of the robot.

For motion planning, a higher-level abstraction of the input space is used.
A sequence of inputs are taken as an \textit{action}, equivalently a \textit{motion primitive}, over a time from $t$ to $t+T$,

\[
\mathbf{a}_t = [\mathbf{u}_t, \mathbf{u}_{t+1}, \hdots, \mathbf{u}_{t+T}].
\]



The transition model from $\mathbf{x}_t$ to $\mathbf{x}_{t+1}$ is taken with respect to actions $a_t$, not sampling time, and is represented as an unknown function

\[
\mathbf{x}_{t+1} = \mathbf{x}_t + F(\mathbf{x}_t, \mathbf{a}_t), \quad \quad F(\mathbf{x},\mathbf{a}^i) = [\Delta x^i, \Delta y^i, 0].
\]

\noindent where each $\Delta x^i$, $\Delta y^i$ are a translation due to action $i$.
However, state transitions are assumed in the planner to change the position of the robot, but not its orientation: $\theta_{t+1} = \theta_t$.
Such an assumption is based on the idea of the ``leading limb'' of the biological brittle star \cite{Astley2012}, where the animal changes direction of movement by making a different choice of ``front" limb''.
Sec. \ref{sec:results} shows that this assumption is reasonable based on hardware data, given that the computer vision system closes the loop by re-sampling $\theta$ between each execution of the planner. Even so, because this choice of leading limb is made at each step, the algorithm is still robust to large deviations from this assumption because each choice is made based only on the current state of the robot.

The ``leading limb'' concept of the brittle star implies five motion primitives: one for each limb which could `lead'.
Therefore, each $\Delta x$, $\Delta y$ is more easily parameterized in polar coordinates, at even intervals of (360/5) $=$ 72 degrees (Fig. \ref{fig:sys_arch}C).
For action $i$,

\[
[\Delta x, \Delta y] = [ r^i \cos(\theta_t + \phi^i), \; r^i \sin(\theta_t + \phi^i) ],
\]

\noindent and the five actions are then each a displacement and an orientation, assuming that leg 0 is pointed along the $+Y$ axis at $90^\circ$,

\[
\begin{bmatrix} r^i \\ \phi^i \end{bmatrix} = \begin{bmatrix} r^0 & r^1 & r^2 & r^3 & r^4 \\ 90^\circ & 18^\circ & 306^\circ & 234^\circ & 162^\circ \end{bmatrix}
\]


The sequence of SMA activations ($\mathbf{u}_t \hdots \mathbf{u}_{t+T}$) for each motion primitive $\mathbf{a}^i$ were hard coded after a trial and error discovery method to determine sequences that enabled linear motion across the plane. 
Multiple gait types were prototyped and one is presented here as the basis for our closed loop locomotion study (details for $\mathbf{u}$ in accompanying software). 
Based again on the `leading limb' model of the biological brittle star, the motion primitive performs a swing-stance gait of the two limbs to each side of the `leader', so that each limb pushes the robot forward simultaneously.
The $r^0 \hdots r^4$ are arbitrarily chosen to be 5 cm in the transition model: this allows the planning framework to make predictions based on direction, to a reasonable level of noise on the computer vision system.
Sec. \ref{sec:results} evaluates the true $r^i$ for each of the five leading-limb primitives, data given in Fig. \ref{fig:data_results}B.




\subsection{Motion Planning with Feedback}\label{sec:feedback_control}


Given a hard-coded set of possible actions and predicted transitions, a search-based motion planning algorithm was implemented for PATRICK to follow a greedy policy towards its goal.
The brittlestar$\_$agent node stores a list of possible actions, $\mathbf{a}_t \in \mathcal{A} = \{\mathbf{a}^0, \hdots \mathbf{a}^4\}$ for this 5-action library.
Algorithm \ref{alg:path} is this policy, using feedback in the form of state estimates $\mathbf{x}_t$ from the computer vision system.

The greedy policy itself is a sequential optimization over the set of motion primitives, minimizing the cost function of Euclidean distance to the goal, $\bar{\mathbf{x}}$, as in


\[
c(\mathbf{x}_t) = ||\mathbf{x}_t - \bar{\mathbf{x}}||_2.
\]

\noindent The model for $F(\cdot, \cdot)$ and $\Delta x^i$, $\Delta y^i$ is used by the greedy algorithm to predict next states based on each action.
Algorithm \ref{alg:path} currently uses a single-step horizon.
If extended to multi-step horizon, the mismatch between the predictive model and the actual system would quickly cause the system to diverge from desired behavior, but by reevaluating at every time step, the effects of model error are minimized.  



\begin{algorithm}[htpb]
    
    \SetKwInOut{Input}{Input}
    \SetKwInOut{Output}{Output}

    \Input{Robot state $\mathbf{x}_t$,  goal position $\bar{\mathbf{x}}$, set of primitives $\mathbf{a} \in \mathcal{A} =\{\mathbf{a}^0,\mathbf{a}^1,\mathbf{a}^2,\mathbf{a}^3,\mathbf{a}^4\}$, transition function $\mathbf{x}_{t+1} = \mathbf{x}_t + F(\mathbf{x}_t,\mathbf{a}_t)$, tolerance $d$ for distance to goal}
    \Output{Closed loop trajectory of motion primitives $\mathbf{a}_t$ to get to the goal}
    \While{$||\mathbf{x}_t-\bar{\mathbf{x}}||_2 > d$}
    {
        \For{$\mathbf{a}^i \in \mathcal{A}$}
        {
            ${\hat{\mathbf{x}}_{t+1}} = \mathbf{x}_t + F(\mathbf{x}_t, \mathbf{a}^i)$
            
            $c(\hat{\mathbf{x}}_{t+1}) = ||\hat{\mathbf{x}}_{t+1} - \bar{\mathbf{x}}||_2$
            
        }
        $\mathbf{a}^t =\displaystyle{\argmin_{\mathbf{a}^i}} \; c(\hat{\mathbf{x}}_{t+1})$
        
        $\mathbf{x}_{t+1} \gets \text{Robot.execute}(\mathbf{a}^t)$
    }
    \caption{A greedy model-based policy}\label{alg:path}
\end{algorithm}

\section{RESULTS}\label{sec:results}
\begin{figure}[thpb]
    \centering
    \includegraphics[width=0.5\textwidth]{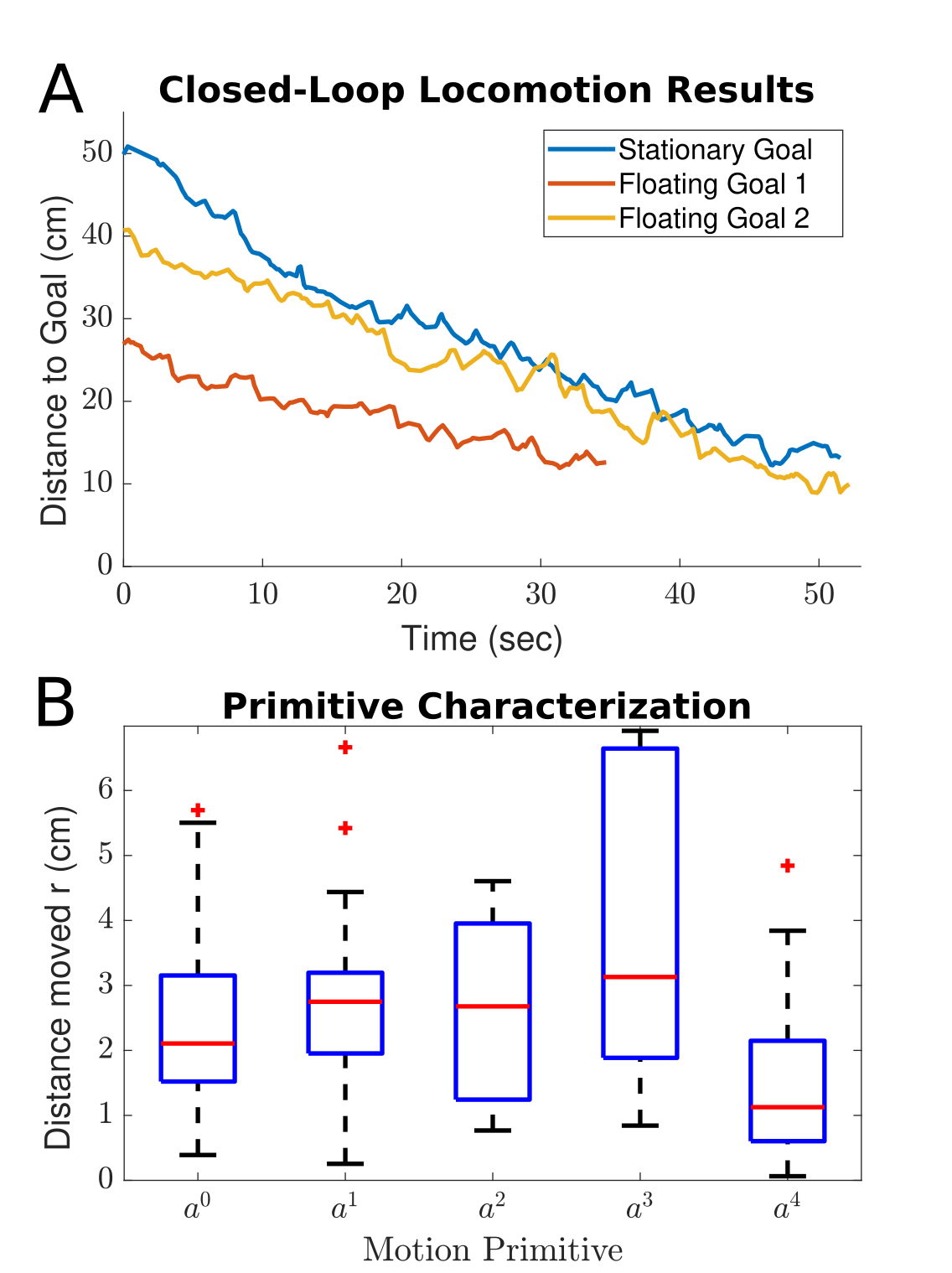}
    \caption{(A) Movement of the robot during three representative hardware tests of the goal-seeking policy, as tracked by the computer vision system. In the ``floating goal'' tests, the target moves over time, and the closed-loop planner adjusts the robot's course. The greedy policy gives an almost-monotonic decrease in cost, even in hardware. (B) Characterization of the actual state transitions induced by executing the primitives. In other words, once a leading limb is chosen - corresponding one to one with the set of primitives - this is the magnitude of the net displacement in the direction of that leading limb. }
    \label{fig:data_results}
\end{figure}

We characterized the execution characteristics of the primitives, illustrated in Fig. \ref{fig:data_results}B.  With the symmetric gait gait presented in this work,  the mean distance the robot covers per iteration is 2.31 cm, and the mean execution time was 2.52 seconds. This works out to approximately 1 centimeter per second ($\sim$0.04 body-lengths/s). 

\begin{figure}[thpb]
    \centering
    \includegraphics[width=0.4\textwidth]{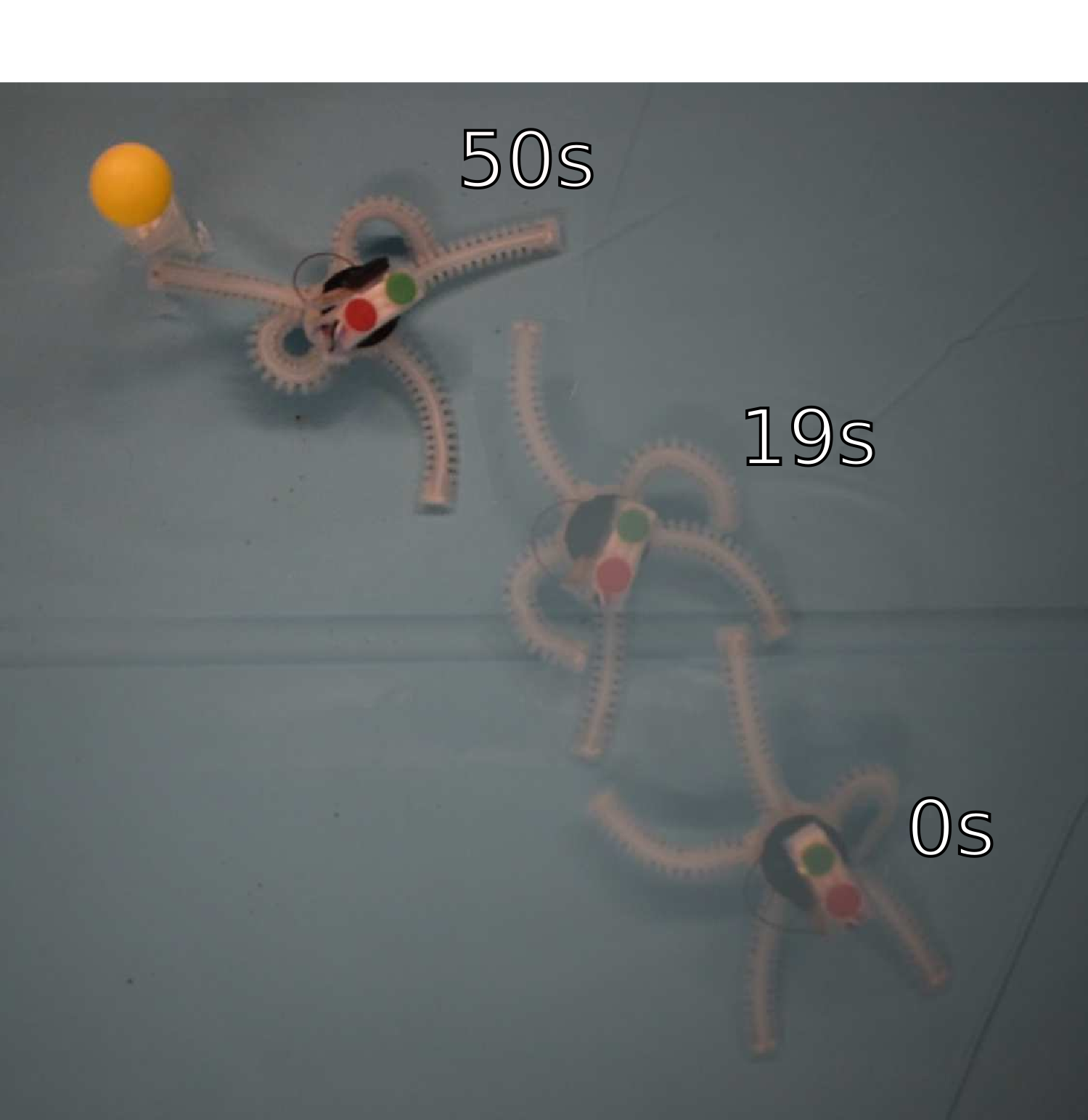}
    \caption{Stationary-goal locomotion test (yellow ball is fixed to the terrain), with frames of the robot along its path to the goal and timestamps for each image. The robot moves roughly 40 cm over the course of 50 seconds to reach the goal state, stopping before collision.}
    \label{fig:robot_test}
\end{figure}

Results from our goal finding experiments are presented in Fig. \ref{fig:data_results}A and an image of the experiment called "Stationary Goal" is shown in Fig. \ref{fig:robot_test}. The robot executes its greedy policy and moves towards the goal with a near monotonic decrease in cost. It is important to note that in all cases the distance to goal never reaches 0 because the robot runs into the goal and its arms - 10 centimeters long - prevent it from getting any closer.

\section{DISCUSSION}
While PATRICK was not optimized for speed, its average velocity of 1 cm/s is comparable to the range of speeds of biological brittle stars (roughly 0.5 - 2 cm/s) \cite{Astley2012}. It achieves this speed using the limited, unparameterized motion primitives characterized in  Fig \ref{fig:data_results}B. It is notable that the distance varies significantly both within each primitive and between them. The inter-limb variance is due to uncharacterized limb force output and displacement which are sensitive to differences in manufacturing. Additionally, because of the complex nature of the SMA actuators coupled with the rubber beam which comprises each limb, every actuator is itself a nonlinear dynamical system with a large amount of hysteresis. While it is tractable to model such a system with a deterministic dynamical systems model \cite{Zakerzadeh2011} \cite{Shu1997}, with DER \cite{Goldberg2019}\cite{Huang2020}, or within the broad framework of geometric mechanics \cite{Renda2018}, the open-loop performance of the system is likely still not sufficient for useful, repeatable behavior. 

Given this variability in limb functionality and primitive execution, it is notable that PATRICK is still able to reach the goal. This means that the strategy of closed-loop planning over high level behaviors enables a mobile robot with deformable elements to deal with a great degree of uncertainty, in line with findings of various other soft robotics groups \cite{Cianchetti2015,Yang2018,Marchese2014} and robotics groups in general \cite{Kalakrishnan2011,Mellinger2010,Johnson2011,Vonasek2013}. For PATRICK, as for others, this expressiveness is possible even in the context of the system's noisy and complicated dynamics due to the deformable structure's robustness to disturbance and the empirical, high-level approach to control. The "morphological intelligence" of these structures obviates many of the potential problems caused by variance in the execution of primitives. For rigid systems, small noise in actuation space can cause large, potentially unsafe variations in the behavior of the physical system. The continuum nature of the limbs of the soft system both reduces discontinuities in the task space evaluation of nominal primitives, as well as increases the safety of executing imprecise primitives. This reinforces the idea that we can produce robots that leverage complex dynamics to successfully navigate in an uncertain world without sacrificing the ability to perform useful high level tasks.

\section{CONCLUSIONS AND FUTURE WORK}
We have presented PATRICK, a soft robotic brittle star and the first untethered underwater soft robot that plans over motion primitives to reach a goal. PATRICK demonstrates the utility of designing soft robots with a high dimensional actuation space, allowing deeper exploration of planning and controls principles. Iteratively planning over high-level motion primitives and leveraging vision-based state feedback enables a great degree of improvement over the open loop behavior and allows the robot to perform useful high level tasks. We believe that, taken together, these features represent a useful contribution in the development of autonomous soft robots, particularly for mobile applications when onboard power and control is necessary.

Future work will focus on broadening the variety of tasks that can be accomplished and developing techniques for controlling the system which take into account the specific implications of nonlinear soft structure and actuation - here, we only scratch the surface of these high level goals. Due to the high dimensionality of PATRICK's actuation space, the limit on interesting behaviors becomes a matter of intelligent planning and control, rather than the limits of the platform itself. For example, using more sophisticated tactile sensing and primitives for grasping, this platform could potentially navigate to a submerged object, pick it up, and move it to another location or otherwise manipulate the object. While this is not a simple task, the physical capabilities of the system are not the limiting factor. The flexibility to serve as both manipulator and self manipulator is an uncommon one in mobile robots, and provides an interesting surface of problems to explore.


There is also much room for improvement and optimization. The limbs can be modeled, characterized, and optimized for a given application. By using statistical learning instead of a trial and error process, we can discover better motion primitives.  Finally, although the use of computer vision for feedback control represents an important contribution of this work, the ultimate goal is to produce a robot that functions independently of an offboard camera. To do that, we need to include sensors on the robot that will enable us to close the loop both inside and outside of lab settings.

\bibliographystyle{IEEEtran}
\bibliography{bib}

\end{document}